\documentclass[graybox]{svmult}

\usepackage{mathptmx}
\usepackage{helvet}
\usepackage{courier}
\usepackage{type1cm}

\usepackage{makeidx}
\usepackage{graphicx}

\usepackage{multicol}
\usepackage[bottom]{footmisc}

\makeindex

\usepackage{float}
\usepackage{multirow}
\usepackage{tabularx}

\begin{document}

\title*{Robust Dialog State Tracking for Large Ontologies}
\author{Franck Dernoncourt, Ji Young Lee, Trung H. Bui, and Hung H. Bui}
\institute{Franck Dernoncourt \at Adobe Research, San Jose, CA, USA, and MIT, USA \quad \email{francky@mit.edu}
\and Ji Young Lee \at Massachusetts Institute of Technology, Cambridge, MA, USA \quad \email{jjylee@mit.edu}
\and Trung H. Bui \at Adobe Research, San Jose, CA, USA \quad \email{bui@adobe.com}
\and Hung H. Bui \at Adobe Research, San Jose, CA, USA \quad \email{hubui@adobe.com}}
\maketitle

\abstract{}

The Dialog State Tracking Challenge 4 (DSTC 4) differentiates itself from the previous three editions as follows: the number of slot-value pairs present in the ontology is much larger, no spoken language understanding output is given, and utterances are labeled at the subdialog level. This paper describes a novel dialog state tracking method designed to work robustly under these conditions, using elaborate string matching, coreference resolution tailored for dialogs and a few other improvements. The method can correctly identify many values that are not explicitly present in the utterance. On the final evaluation, our method came in first among 7 competing teams and 24 entries. The F1-score achieved by our method was 9 and 7 percentage points higher than that of the runner-up for the utterance-level evaluation and for the subdialog-level evaluation, respectively.

\section{Introduction}
\label{sec:introduction}

Spoken dialog systems are computer-based systems that interact with users through natural language to help them achieve a goal. Since they provide a convenient and natural way for the users to interact with computers, they have become increasingly popular recently. Examples of such systems are Apple Siri, Google Now, Microsoft Cortana, and Amazon Echo. 

A dialog state tracker is a key component of a spoken dialog system and its goal is to maintain the dialog states throughout a dialog. For example, in a tourist information system, the dialog state may indicate a tourist attraction that the user is interested in, as well as the types of information related to the attraction, such as entrance fees and location. Although it might be easy for humans, dialog state tracking is difficult for computers because they do not ``understand'' natural human language as humans do. Yet, dialog state tracking is crucial for reliable operations of a spoken dialog system because the latter relies on the estimated dialog state to choose an appropriate response, for example, which tourist attractions to suggest.

The Dialog State Tracking Challenge (DSTC) is a series of community challenges that allow researchers to study the state tracking problem on a common corpus of dialogs  using the same evaluation methods. DSTC 4, which is the focus of this paper, differentiates itself from previous editions in several ways. First, the ontology contains many more slot-value pair: for example, the ontology for the DSTC 3 corpus contains only 9 slots and 271 values, while DSTC 4's ontology has 30 slots and 1667 values, most of which are shared across several slots. Second, the output of the Spoken Language Understanding (SLU) component, is not available. Third, the states are labeled at the subdialog level only, which disadvantages machine-learning-based approaches.

Best approaches in the previous DSTCs include neural networks \cite{henderson2013deep, henderson2014robust, henderson2014word}, web-style ranking and SLU combination \cite{williams2014web}, maximum entropy models~\cite{lee2013recipe} and conditional random fields \cite{ren2013dialog}. However, these prior solutions are limited to domain ontologies with a small number of slots and values. Furthermore, most of the above methods and other rule-based methods \cite{sun2014generalized} require the output of the SLU.

This paper proposes a novel dialog state tracking method designed to work robustly under the DSTC 4 conditions. The method is composed of an elaborate string matching system, a coreference resolution mechanism tailored for dialogs, and a few other improvements. The paper is organized as follows. Section 2 describes briefly the DSTC 4 data set. Section 3 presents in detail several trackers we used for the challenge. Section 4 compares the performances on the test set of our trackers with those of the trackers submitted by other teams participated in the challenge. Section 5 summarizes our work and proposes further improvements. 
\section{The DSTC 4 data set}
\label{sec:dstc}

The corpus used in this challenge consists of 35 dialog sessions on touristic information for Singapore, collected from Skype calls between three tour guides and 35 tourists. Each dialog session is a dialog between a guide and a tourist, where the guide helps the tourist plan for a trip to Singapore by providing recommendations based on the tourist's preferences. These 35 dialogs sum up to 31,034 utterances and 273,580 words. All the recorded dialogs with the total length of 21 hours have been manually transcribed and annotated with speech act and semantic labels for each utterance as well as dialog states for each subdialog. 

Each dialog is divided into subdialogs, each of which has one topic and contains one or several utterances. Dialog states are annotated for each subdialog. A dialog state is represented as a list of slot-value pairs. The slot is a general category, while the value indicates more specifically what the dialog participants have in mind. For example, one possible slot is ``TYPE OF PLACE'' and a possible value of this slot is ``Beach''. The DSTC 4 corpus is provided with an ontology that specifies the list of slot-value pairs that a subdialog of a given topic may contain.

Following the official split, the train, development and test sets contain 14, 6 and 9 dialogs respectively. The remaining 6 dialogs are used as a test set for another task. The test set labels were released only after the final evaluation.

The goal of the main task of DSTC4 is to track dialog states, considering all dialog history up to and including the utterance. Trackers are evaluated based on the predicted state for either each utterance (utterance-level evaluation) or for each subdialog (subdialog-level evaluation). Since the gold labels are available only at the subdialog level, in the utterance-level evaluation the predicted state for each utterance is compared against the gold labels of the subdialog that contains the utterance. Four performance metrics are used: subset accuracy, precision, recall and F1-score. For the subset accuracy, for a given utterance, the list of all slot-value pairs in the dialog state must exactly match the subdialog gold label to be counted as a true positive. \cite{DSTC4handbook, DSTC4} contain further information pertaining to the data set.

\section{Method}
\label{sec:method}

This section presents the dialog state trackers we used for the challenge. We describe two rule-based trackers, two machine-learning-based trackers and a hybrid tracker.

\subsection{Fuzzy matching baseline}
\label{sec:fuzzy-matching}

A simple rule-based tracker was provided by the organizers of the challenge. It performs string fuzzy matching between each value in the ontology and the utterance. If the matching score is above a certain threshold for a value, then any slot-value pair with that value is considered as present.

\subsection{Machine-learning-based trackers}
\label{sec:ml-trackers}

\subsubsection{Cascade tracker}
\label{sec:cascade-tracker}
The cascade tracker aims to refine the fuzzy matching tracker. For each slot, a classifier is trained to detect whether the slot is present or absent given an utterance. If a slot is predicted as present for a given utterance, then the fuzzy matching score is computed between each value of the detected slot and the utterance. For the classifier, we tried logistic regression (LR), support vector machines (SVM), and random forests (RF): we kept RF as it yields the best results. The features used are unigrams, bigrams, word vectors and named-entities. The word vector features are obtained by first mapping each word of the utterance to a word vector, then summing them all. We use pre-trained word vectors provided on the word2vec website\footnote{\url{https://code.google.com/p/word2vec/}: GoogleNews-vectors-negative300.bin}.

\subsubsection{Joint tracker}
\label{sec:joint-tracker}
The main weakness of the cascade tracker is that in order to detect the value, it relies on fuzzy matching instead of utilizing more meaningful features.
To address this issue, the joint tracker predicts the slot and the value jointly. For each slot, an RF classifier is trained to detect whether a given value is present or absent. The features used are the same as in the cascade tracker. Since the vast majority of values are absent in a given utterance, the negative examples are downsampled in the training phase.

\subsection{Elaborate rule-based tracker}
\label{sec:elaborate}

\begin{figure}[H]
\sidecaption
\includegraphics[width=\linewidth]{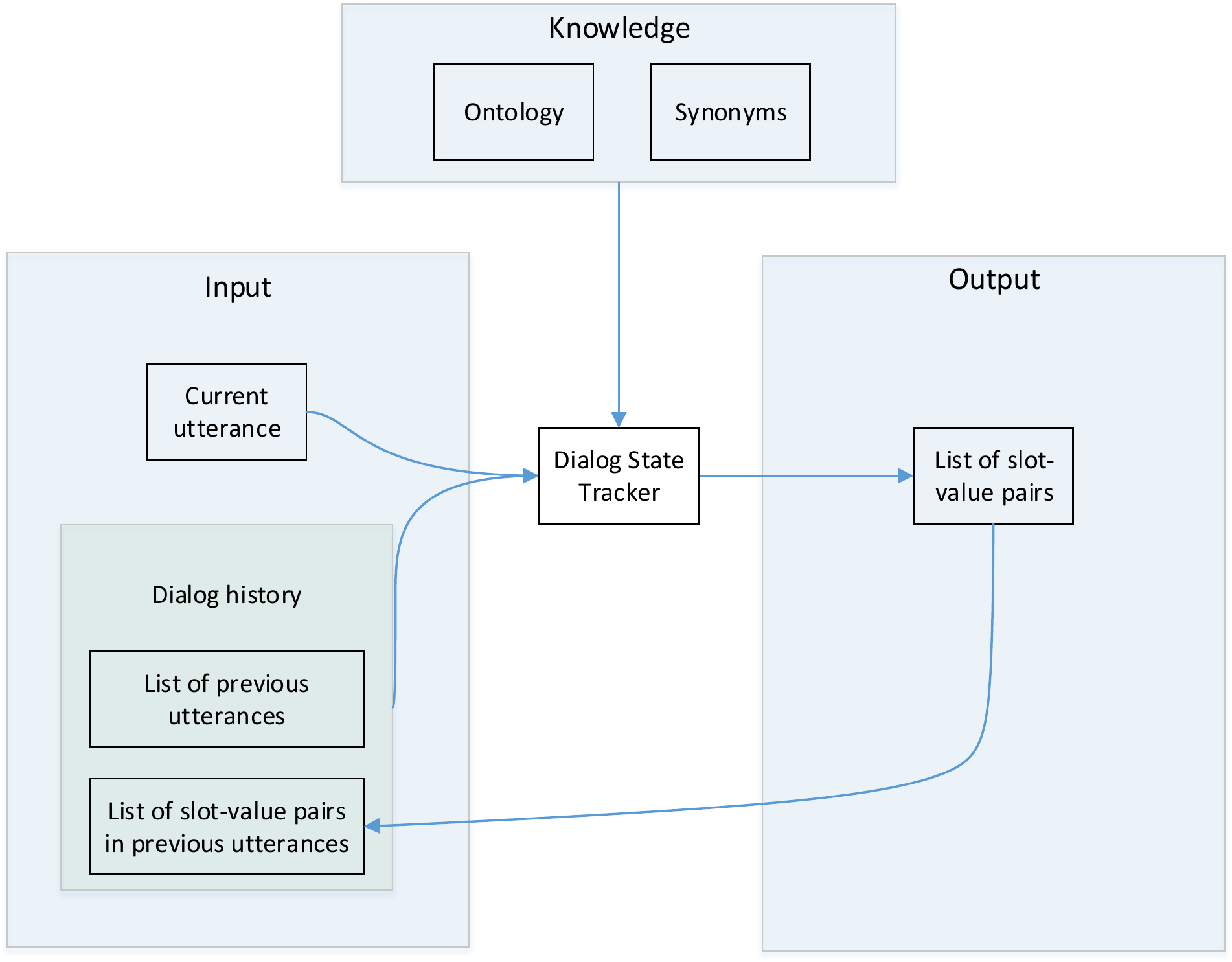}
\caption{Overview of the dialog state tracking system.}
\label{fig:tracker-diagram}
\end{figure}

Since the machine-learning-based approaches using traditional features were performing poorly, an elaborate rule-based tracker was constructed in order to overcome the shortcomings of the machine-learning-based approaches. 
The main pipeline of the elaborate rule-based tracker is described in Figure~\ref{fig:tracker-diagram}. The tracker makes use of the knowledge present in the ontology as well as the synonym list that is defined for each slot-value pair.
The inputs of the dialog state tracker are the current utterance (i.e. the utterance for which the tracker should predict the slot-value pairs), and the dialog history.
The dialog history contains the list of previous utterances, as well as the list of slot-value pairs that the tracker predicted for the previous utterances.
Lastly, based on the input and the knowledge, the tracker outputs a list of slot-value pairs for the current utterances.

\begin{figure}[H]
\sidecaption
\includegraphics[width=\linewidth]{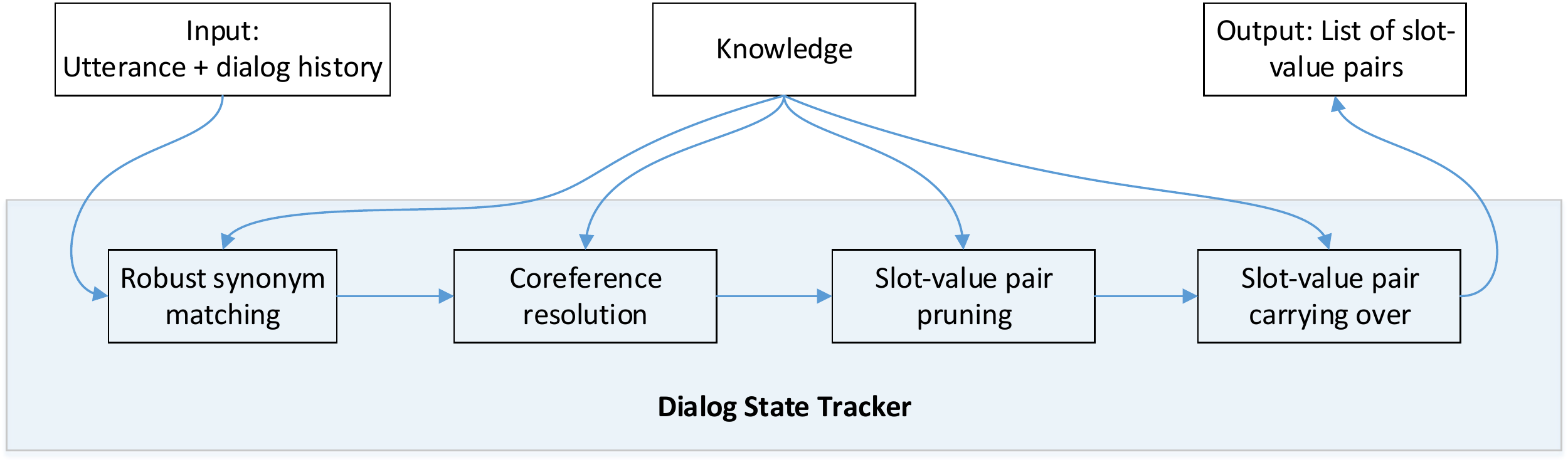}
\caption{The four main steps of the elaborate rule-based tracker.}
\label{fig:tracker001}
\end{figure}

This tracker tries to model how a human would track the dialog states, and therefore is very intuitive and interpretable. Figure~\ref{fig:tracker001} presents the four main steps the tracker follows to predict an output based on the input and the knowledge. The first step detects the presence of each slot-value pair in the utterance, by finding a match between any substring of the utterance and any of the synonyms of each slot-value pair. The second step resolve coreferences of certain type and detects additional slot-value pairs associated with them. Among the slot-value pairs detected from synonym matching and coreference resolution, there often exist a group of slot-value pairs that are closely related, but only one of them is present in the dialog state. In the third step, the tracker selects the most likely slot-value pair among the closely-related slot-value pairs and eliminates all others. In the last step, slot-value pairs are carried over from the previous dialog state, whenever the topic continues and no new slot-value pair is detected for certain slots. The following four subsections present each step in more details.

\subsubsection{Robust synonym matching}
\label{subsec:synonym matching}

The motivation behind the synonym matching comes from the observation that even when a slot-value pair is included in the dialog state of an utterance, the value itself does not appear in the utterance. Instead, a synonym of the value often appears as a substring of the utterance. For example, for the slot-value pair ``PLACE: Amoy by Far East Hospitality 4'', it is likely that a synonym of the value such as ``Amoy Hotel'' is present in an utterance of a spoken dialog, rather than the value itself, viz. ``Amoy by Far East Hospitality 4''. Therefore, each slot-value pair is associated with a list of synonyms that are likely to be present in the utterances whose dialog state contains the slot-value pair. The synonym list was created partly by hand and partly by using a set of rules. 

For flexibility and better detection, each synonym may contain two optional specifications: first, a synonym could be expressed as an AND clause of multiple words; second, part-of-speech (verb or noun) may be specified for each word that comprise a synonym. If a synonym is an AND clause of multiple words, the corresponding slot-value pair is detected only when each of the words is present in the utterance. This allows a slot-value pair to be detected even when the word order is switched. For example, for the slot-value pair ``PLACE: Amoy by Far East Hospitality 4'' discussed earlier, if a synonym (``Amoy'' AND ``Hotel'') is added to the corresponding synonym list, then the slot-value pair will be detected in an utterance such as ``I recommend the hotel called Amoy.'' 

If a word that comprises a synonym is specified with a part-of-speech, then the corresponding slot-value pair is detected only when the word appears in the utterance and is tagged as having the specified part-of-speech. To take into account for the part-of-speech, the tracker performs part-of-speech tagging on each utterance prior to the synonym matching. The part-of-speech specification allows a slot-value pair to be detected even when a synonym word is exhibited in a different form. For example, if we use a synonym ``Snorkel'' with verb as the part-of-speech for the slot-value pair ``ACTIVITY: Snorkeling'', then the slot-value pair will be detected in the utterances ``I like to snorkel'', ``Have you snorkeled before?'', and ``There are many people snorkeling in this beach''. Another benefit of using part-of-speech specification is that it helps reduce many false positives by distinguishing between noun and verb instances of a word. For example, specifying noun as the part of speech in a synonym ``Show'' for the slot-value pair ``ACTIVITY: Show'' will prevent the incorrect detection of the slot-value pair in an utterance such as ``I would like to show you this picture.''

Moreover, we adopted two simple strategies to enhance the detection of slot-value pairs. First, we lemmatized each word in both the synonyms and the utterances before matching, increasing the chance of detecting the plural as well as singular form of a synonym. Second, in order to account for misspellings while preserving the precision of the tracker, we permitted one spelling mistake on long synonyms only. Specifically, we allowed a synonym to be detected if a substring of the utterance had Levenshtein distance of 1 from the synonym, only when a synonym has more than 5 characters and each word in a synonym has more than 3 characters.

\subsubsection{Coreference resolution for dialogs}

Coreferences are numerous in dialogs but are harder to detect than in formal written text, as the existing coreference resolution systems typically perform well on the latter, but not on the former. 

The tracker contains a coreference resolution system for place-related anaphoras. This system is customized for the slot-filling tasks and works as follows. For each utterance, from its syntactic parsing tree the tracker detects the presence of the following three templates:
\begin{itemize}
\item Template 1: possessive adjective (my/your/our) + a type of place
\item Template 2: demonstrative pronoun (the/this/that/these/those) + a type of place
\item Template 3: here/there
\end{itemize}
For example, ``our hotel''and ``your museums'' belong to Template 1, and ``this garden'' and ``these parks'' belong to Template 2.
If Template 1 or Template 2 is present, then the tracker considers as present the last detected slot-value pair of the same type in the dialog history. The type of each place-related slot-value pair is specified in the provided ontology.
If Template 3 is present, then the tracker considers as present the last detected value of any place-related slots (e.g. ``PLACE'' or ``NEIGHBOURHOOD'') in the dialog history.

\subsubsection{Ontology-based slot-value pair pruning}
Among the slot-value pairs detected from synonym matching and coreference resolution, there often exist a group of slot-value pairs that are closely related, e.g. different branches of the same hotel chain.
In most situations, however, only one of these slot-value pairs is present in the dialog state.
In order to select the most likely slot-value pair among the closely related slot-value pairs, the tracker utilizes the domain knowledge present in the ontology as well as the observations from the training data. 

For example, for each hotel listed as a possible value, the ontology also contains additional information about the hotel such as its neighborhood and price range. When multiple hotel branches are detected from the synonym matching step, then the tracker checks whether other related information about the branch is found in the context and selects the most likely branch based on the observation. If no relevant information is found, the tracker selects the most likely branch based on prior observations from the training data.     

Another kind of closely-related slot-value pairs are those with the values that overlap with each other, such as ``Park Hotel'' and ``Grand Park Hotel''. If the utterance is ``I will stay at the Grand Park Hotel'', then the synonym matching step will detect both ``Park Hotel'' and ``Grand Park Hotel'' values. To avoid this issue, the tracker deletes any slot-value pair whose value is a (strict) substring of the value of another slot-value pair, among the detected slot-value pairs. 

For slot-value pairs of special slots such as ``TO'' and ``FROM'', syntactic parsing trees are used in order to determine whether each value follows a preposition such as ``to'', ``into'', ``towards'', and ``from''. Based on this and the order in which the values appear in the utterances of a subdialog, the most likely slot-value pair(s) are determined.

\subsubsection{Slot-value pair carrying over}

As a dialog progresses, many slot-value pairs remain present for several subsequent utterances as the subject of the dialog continues. As a result, the tracker implements the following rule: for certain slots, whenever a slot-value is detected as present in the previous utterance, the slot-value pair remains present until another value appears for the same slot or the topic changes. The tracker learns for which slots it is optimal to do so by using the training data and comparing the slot-value pair prediction results with and without the rule for a given slot.

\subsection{Hybrid tracker}
\label{sec:hybrid-based}

\begin{figure}[H]
\sidecaption
\includegraphics[width=\linewidth]{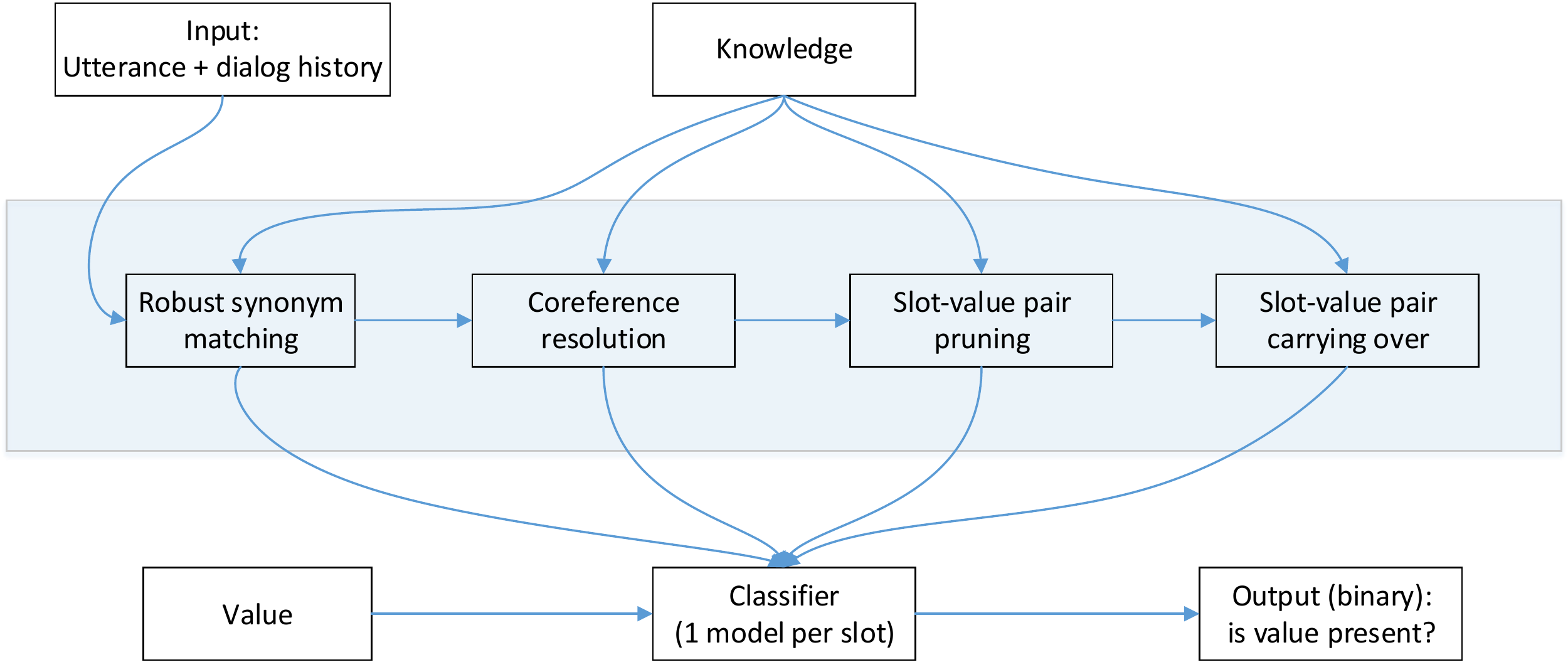}
\caption{The hybrid tracker uses the output of each of the four main steps of the elaborate rule-based tracker as features.}
\label{fig:RB-tracker-as-LR-input}
\end{figure}

In order to take advantage of the strength of both the rule-based and the machine-learning-based approaches, the hybrid tracker uses the rule-based tracker's outputs as features for the joint tracker. The output of each of the four main steps of the elaborate rule-based tracker is used as features, as Figure~\ref{fig:RB-tracker-as-LR-input} illustrates.

\section{Results}
\label{sec:results}

Table~\ref{tab:results} compares the performances on the test set of our trackers as well as the best tracker of each other team that took part in the challenge. For the utterance-level evaluation, all teams but two including us obtained an F1-score below 0.35, which reflects the difficulty of the task. Team 4 reached 0.4481, while our best entry scored 0.5306. By the same token, for the subdialog-level evaluation, all teams but two including us obtained an F1-score below 0.40. Team 4 reached 0.5031, while our best entry scored 0.5786.

Looking at the results for various trackers described in Section \ref{sec:method}, we observe that the cascade tracker and the joint tracker both perform poorly. The joint tracker has a much lower recall than the cascade tracker, which may be due to the fact that the same classifier is used for all values of a slot. However, the elaborate rule-based tracker yields a much higher performance, far above the fuzzy matching baseline. The hybrid tracker, which uses the output of the rule-based tracker as features, further increases the F1-score, but has a lower subset accuracy. Unlike the joint tracker, the hybrid tracker improves the F1-score, which may result from the higher quality and density of the features used.

\newcolumntype{?}{!{\vrule width 1pt}}
\newcolumntype{C}{>{\centering\arraybackslash}X}%
\renewcommand*\arraystretch{1.1}
\begin{table}[h]
\caption{Comparison of results for various dialog state trackers on the test set}
\label{tab:results}
\begin{tabularx}{\textwidth}{X|CCCC|CCCC}
\svhline
\multirow{2}{*}{ Tracker  } & \multicolumn{4}{c|}{Utterance-level} & \multicolumn{4}{c}{Subdialog-level} \\
 & Accuracy & Precision & Recall & F1-score & Accuracy & Precision & Recall & F1-score  \\%\vspace{0.1cm}\\

\hline
Baseline	& 0.0374	& 0.3589	& 0.1925	& 0.2506	& 0.0488	& 0.3750	& 0.2519	& 0.3014 \\
Cascade		& 0.0227	& 0.2962	& 0.2145	& 0.2488	& 0.0314	& 0.3138	& 0.2734	& 0.2922 \\  
Joint		& 0.0260	& 0.4682 	& 0.1170	& 0.1872	&  0.0357	&  0.4648	&   0.1602	& 0.2383 \\	 
Elaborate	& \textbf{0.1210}	& 0.5449	& \textbf{0.4964}	& 0.5196	& \textbf{0.1500}	& 0.5619	& \textbf{0.5787}	& 0.5702 \\
Hybrid		& 0.1183	& \textbf{0.5780}	& 0.4904	& \textbf{0.5306}	& 0.1473	& \textbf{0.5898}	& 0.5678	& \textbf{0.5786} \\
\hline
Team 4  	& 0.1002	& 0.5545	& 0.3760	& 0.4481	& 0.1212	& 0.5642	& 0.4540	& 0.5031 \\
Team 2  	& 0.0489	& 0.4440	& 0.2703	& 0.3361	& 0.0697	& 0.4634	& 0.3335	& 0.3878 \\
Team 6 		& 0.0486	& 0.5623	& 0.2314	& 0.3279	& 0.0645	& \textbf{0.5941}	& 0.2850	& 0.3852 \\
Team 1 		& 0.0371	& 0.4179	& 0.2804	& 0.3356	& 0.0584	& 0.4384	& 0.3426	& 0.3846 \\
Team 5 		& 0.0268	& 0.3405	& 0.2014	& 0.2531	& 0.0401	& 0.3584	& 0.2632	& 0.3035 \\
Team 7 		& 0.0286	& 0.2768	& 0.1826	& 0.2200	& 0.0323	& 0.3054	& 0.2410	& 0.2694 \\
\svhline
\end{tabularx}
\end{table}

The results for the utterance-level evaluation are lower than for the subdialog-level evaluation, which is expected since for the utterance-level evaluation the predicted state for each utterance is compared to the gold state of the subdialog that contains the utterance. It is often hard or impossible in the first utterances of a subdialog to guess what state the subdialog (i.e. the state of the last utterance of the subdialog) will have, since the tracker is allowed to access only the current and previous utterances, but not the upcoming utterances.

The poor performances of trackers that solely rely on machine-learning can be partly explained by the lack of dialog state labels at the utterance level. If a tracker is trained using the features extracted for each utterance and the subdialog label as the utterance label, then the tracker will learn many incorrect associations between features and labels. In other words, using subdialog labels as utterance labels introduce much noise. 

For example, if a subdialog comprises the two utterances ``Good morning'' and ``Hi! I am on my way to Paris!'', and the subdialog label is ``TO: Paris'', the first training sample will be the features extracted from ``Good morning'' and the label ``TO: Paris''. This will tend to create false positives, since it is likely that ``Good morning'' in other subdialogs will not be labeled with ``TO: Paris''.

However, if a tracker is trained only for subdialogs to avoid this issue, then it results in having much fewer samples to train on. This brings us to the issue of data scarcity: even though the train set contains 14 labeled dialogs, the number of training samples is still quite small. This certainly gives a significant advantage to hybrid trackers over machine-learning-based trackers.

In addition to the dialog states, the labels also contain semantic tags for each utterance. We tried to take advantage of the finer granularity of the semantic tagging: as an experiment, we used the gold semantic tags as features, but our results on the development set did not show any improvement.

\section{Conclusion and future work}
\label{sec:conclusion}

This paper describes and compares several dialog state trackers on the DSTC~4 corpus. Due to the size of the ontology and the utterances being labeled at the subdialogue-level only, the rule-based approach yields better results than the pure machine learning approaches. However, using the rule-based tracker as features for the machine-learning-based tracker allows to further improve the results.
On the final evaluation of the main task, our method came in first among 7 competing teams and 24 entries. Our method achieved an F1-score 9 and 7 percentage points higher than the runner-up for the utterance-level evaluation and for the subdialog-level evaluation, respectively.

Modeled after how humans would track dialog states, our elaborate rule-based tracker is not only intuitive and interpretable, but also has potential to be further improved by combining machine-learning-based approaches. One such example is our hybrid tracker, but there are many other ways that machine learning techniques could be used to improve our system. 

First, the synonyms list was mostly manually curated, as using existing synonym lists such as Wordnet was causing the precision to decrease significantly. Some general rules were used to automatically generate the synonyms, but one could further automate the generation of synonym list. Moreover, extending the coreference resolution system to general slot-value pairs can improve the performance of the tracker. Furthermore, instead of blindly carrying over slot-value pairs when no new value is detected for certain slots, it would be interesting to implement algorithms that can detect when the subject of the dialog has changed and only carry over slot-value pairs when the subject has not changed.

Another weakness of our system is that it detects all slot-value pairs that are mentioned in the utterances, rather than selectively detecting those that are not only mentioned, but also the main subject of discussion. One example is when a value is mentioned but negated, e.g. ``I recommend Keong Saik Hotel, not The Fullerton Hotel''. Then according to our system both slot-value pairs ``PLACE: Keong Saik Hotel'' and ``PLACE: The Fullerton Hotel'' will be detected as present, but the gold dialog state will only include the former. Such mistakes result in many false positives. Implementing algorithms to detect the main subject may greatly improve the precision.

\begin{acknowledgement}
The authors would like to warmly thank the DSTC 4 team for organizing the challenge and being so prompt to respond to emails. The authors are also grateful to the anonymous reviewers as well as to Walter Chang for their valuable feedback.
\end{acknowledgement}
\bibliographystyle{abbrv}
\bibliography{iwsds2016}

\end{document}